\documentclass[10pt,letterpaper]{article}

\usepackage[hidelinks]{hyperref}

\usepackage[table]{xcolor} 
\usepackage{lipsum}
\usepackage{graphicx} 
\usepackage{tikz} 
\usepackage{footnote}
\usepackage{threeparttable}  
\usepackage{tcolorbox} 
\usepackage{amsthm} 
\usepackage{amsmath} 
\usepackage{array} 
\usepackage{enumitem} 
\usepackage{geometry} 
\usepackage{longtable} 
\usepackage{tabularx} 
\usepackage{booktabs} 
\usepackage{multirow} 
\usepackage{datetime} 
\usepackage{pslatex} 
\usepackage{apacite} 
\usepackage{lineno} 
\usepackage{tikz}
\usepackage{amsmath}
\usepackage{array}
\usepackage{graphicx}
\usepackage{mathabx}  
\usepackage{ragged2e}

\usepackage{xr}
\usepackage{url}
\newcommand{\textdataset}{TextualChoices-1K}
\newcommand{\petersondataset}{Choices13k}
\newcommand{\ExampleA}{This option may seem appealing for its consistency, but it cannot offer any surprisingly high rewards}
\newcommand{\ExampleB}{This alternative holds an advantage for the risk-takers who seek the excitement of a larger possible gain}

\usetikzlibrary{positioning, arrows.meta} 

\usepackage{subcaption} 

\usepackage{cogsci}

\usepackage{float} 

\title{Predicting Human Choice Between Textually Described Lotteries}
\cogscifinalcopy 

\author{{\large \bfseries  Eyal Marantz (Eyalmarantz@campus.technion.ac.il)} \\
  Faculty of Data and Decision Sciences\\
  Technion - Israel Institute of Technology, Haifa, 3200003, Israel
  \AND {\large \bf Ori Plonsky (Plonsky@technion.ac.il)} \\
  Faculty of Data and Decision Sciences\\
  Technion - Israel Institute of Technology, Haifa, 3200003, Israel}

\begin{document}

\maketitle

\begin{abstract}

Predicting human decision-making under risk and uncertainty is a long-standing challenge in cognitive science, economics, and AI. While prior research has focused on numerically described lotteries, real-world decisions often rely on textual descriptions. This study conducts the first large-scale exploration of human decision-making in such tasks using a large dataset of one-shot binary choices between textually described lotteries. We evaluate multiple computational approaches, including fine-tuning Large Language Models (LLMs), leveraging embeddings, and integrating behavioral theories of choice under risk. Our results show that fine-tuned LLMs, specifically  GPT-4o, outperform hybrid models that incorporate behavioral theory, challenging established methods in numerical settings. These findings highlight fundamental differences in how textual and numerical information influence decision-making and underscore the need for new modeling strategies to bridge this gap.


\end{abstract}

\section{Introduction}

Predicting and understanding human choice under uncertainty is a fundamental challenge in economics, psychology, and the cognitive sciences, with clear implications for many real-world scenarios, including financial investments, health-related choices, and risk management. Most of the systematic study in this domain has focused on investigating how people choose between lotteries or gambles, with these lotteries explicitly and accurately described using numerical format. This line of research, which goes back more than eight decades, assumes that the response to these numerical descriptions captures the basic properties of human decision making under risk and uncertainty. Therefore, the insights gained in these studies should generalize to more natural settings. Importantly, many of the most important insights such research reveals concern the ways by which people seem to deviate from clear theoretical benchmarks like maximization of Expected Value or of Expected Utility. It is convenient that the numerical format of presentation thus allows computing the predictions of these benchmarks. 

Yet, in the real world, people rarely face precise numerical descriptions of choice options. Instead, potential options may often be described using natural language. For example, people may face signs that warn against choosing certain options or ads that promote the choice of other options. That is, in many real-world situations, rather than relying on precise numerical information, individuals must rely on qualitative descriptions and make subjective interpretations of textual information before reaching a decision. In this paper, we investigate---and try to predict---people's decisions between textually described choice options that do not contain precise numerical information.

\begin{figure}[h]
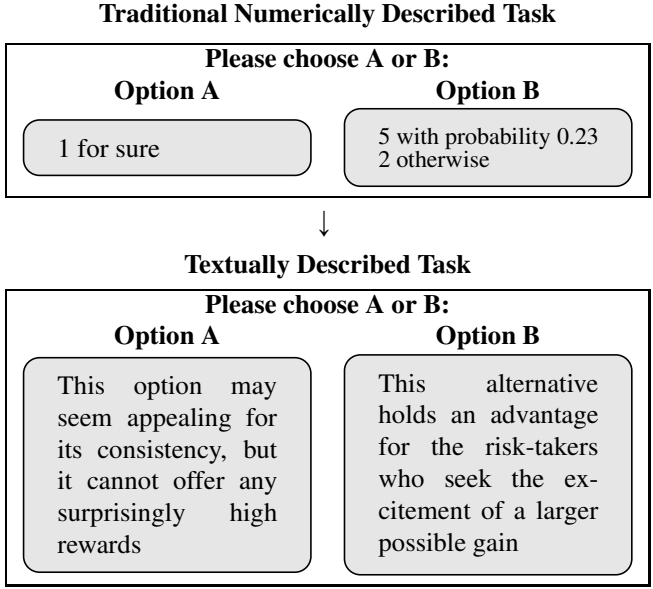

    \centering
    \textbf{Traditional Numerically Described Task} \\[0.5em]

    \begin{tabular}{|m{3.8cm}m{3.8cm}|}  
    \hline
    \multicolumn{2}{|c|}{\textbf{Please choose A or B:}} \\ 

    \multicolumn{1}{|c}{\textbf{Option A}} & 
    \multicolumn{1}{c|}{\textbf{Option B}}  \\ 

    \begin{minipage}[t]{3.8cm}
        \centering
        \begin{tcolorbox}[colframe=black, colback=gray!20, arc=2mm, boxrule=0.6pt, boxsep=1pt]
            1 for sure
        \end{tcolorbox}
    \end{minipage}
    & 
    \begin{minipage}[t]{3.8cm}
        \centering
        \begin{tcolorbox}[colframe=black, colback=gray!20, arc=2mm, boxrule=0.6pt, boxsep=1pt]
            \small \mbox{5 with probability 0.23} \\ \small 2 otherwise
        \end{tcolorbox}
    \end{minipage} \\ \hline
    \end{tabular}

    \vspace{0.4em}
    {\large$\downarrow$}
    \vspace{0.4em}

    \textbf{Textually Described Task} \\[0.3em]
    \begin{tabular}{|m{3.8cm}m{3.8cm}|}  
    \hline
    \multicolumn{2}{|c|}{\textbf{Please choose A or B:}} \\ 

    \multicolumn{1}{|c}{\textbf{Option A}} & 
    \multicolumn{1}{c|}{\textbf{Option B}}  \\ 

    \begin{minipage}[t]{3.8cm}
        \centering
        \begin{tcolorbox}[colframe=black, colback=gray!20, arc=2mm, boxrule=0.6pt, boxsep=1pt]
            \ExampleA
        \end{tcolorbox}
    \end{minipage}
    & 
    \begin{minipage}[t]{3.8cm}
        \centering
        \begin{tcolorbox}[colframe=black, colback=gray!20, arc=2mm, boxrule=0.6pt, boxsep=1pt]
            \ExampleB
        \end{tcolorbox}
    \end{minipage} \\ \hline
    \end{tabular}

    \caption{Comparison of Numerical and Textual Task Descriptions}
    \label{fig:comparison_AB}
\end{figure}

Under a textual description format, almost any behavior may be considered ``rational" (i.e., adhering to the prescriptions of expected value or utility maximization). For example, Figure \ref{fig:comparison_AB} presents a binary choice task presented in two formats. Under a numerical format, the task has a clear theoretical prediction: Option B that provides ``5 with probability .23; 2 otherwise" dominates---and should be chosen over---Option A that provides ``1 for sure". Yet, when described textually, this no longer holds. While the textual descriptions are accurate (in the sense that they faithfully describe the underlying payoff distributions), the choice of Option A (\textit{\ExampleA}) over B (\textit{\ExampleB}) is quite reasonable and depends on both subjective interpretations of the texts and on idiosyncratic preferences. Under the textual format, it is also quite hard to elicit clear predictions of extant computational models of choice that lack the ability to process the textual inputs.

Lacking clear benchmarks, we chose to start the investigation of this domain with a prediction-based study. Using a recently collected dataset of 1000 one-shot binary choice tasks, \textbf{\textdataset{}} \cite{erevinprep} we conduct the first large-scale exploration of human decision-making in tasks framed through textual descriptions, rather than numeric lotteries. We systematically test various computational approaches, all of which use Large Language Models (LLMs) that can accept the textual descriptions as input. Our study contrasts and compares different ways to use LLMs to predict behavior in this task, including both purely data-driven methods and approaches that aim to enhance the predictive ability of the LLMs with extant behavioral theories of choice under risk and uncertainty. In so doing, we also aim to bridge the gap between extant numeric-focused models and modern language-based decision frameworks, advancing our understanding of human decision-making under uncertainty while highlighting the strengths and limitations of LLMs in this context. 

Recent works on predicting numerically described tasks revealed that hybrid methods that complement data-driven computational methods with behavioral theories lead to the most accurate models of choice prediction \cite{plonsky2019predicting}. In contrast, our findings revealed that behavioral-theory-free machine learning models outperform theory-driven models in predicting decisions based on textual descriptions. This has led us to test similar data-driven methods, namely fine-tuning of LLMs, in numerically described tasks. Our results suggest hybrids of behavioral theory and machine learning still outperform the pure LLM approach in these settings.

This divergence may hint of a fundamental difference in the choice processes involved in numerically vs. textually described options. While modern computational models excel at interpreting and predicting decisions based on natural language cues, they face challenges when precision and numeric reasoning are required. These findings cast doubt on the assumptions underlying much of the classical behavioral research on choice under risk and uncertainty and underscore the need for task-specific strategies in computational modeling, tailoring predictive approaches to the structure of the decision problem.

\subsubsection{Related Work} 

Our work is related mainly to two lines of research. First, it relates to studies that aim to predict human decision-making using behavioral models, ML, or a combination of both. Historically, models such as Expected Utility Theory assumed that individuals make decisions by maximizing utility. However, decades of empirical research have shown systematic deviations from this rational framework. This led to the development of behavioral models like Prospect Theory \cite{kahneman1979prospect} and many others, including Best Estimate and Sampling Tools (BEAST) that has shown high accuracy in predicting choice under risk uncertainty \cite{erev2017anomalies}. 

More recently, machine learning (ML) techniques have been combined with behavioral theories to create hybrid models that improve predictive accuracy \cite{peterson2021using,plonsky2017psychological}. For example, BEAST-GB \cite{plonsky2019predicting}, integrates behavioral insights based on the model BEAST with ML tools to achieve state-of-the-art predictive performance for choice between numerically described choice options. While these approaches have advanced decision-making research in settings that involve explicit numeric outcomes and probabilities, they primarily focus on such structured scenarios, leaving a gap in understanding decision-making in less structured, real-world tasks.

Second, our work is related to recent studies that use LLMs to mimic, augment, or predict human behavior. Advances in LLMs have demonstrated their capacity to process and interpret qualitative information effectively. For example, CENTaUR \cite{binz2023turninglargelanguagemodels, binz2024centaur} and Arithmetic-GPT \cite{zhu2024languagemodelstrainedarithmetic} have shown that LLMs can accurately predict human decisions in numeric and arithmetic contexts. However, challenges remain, as LLMs often default to overly rational behavior and struggle with inconsistencies in reasoning \cite{liu2024largelanguagemodelsassume, macmillan2024ir}. Our work examines the usefulness of LLMs for prediction of human choice when clear benchmarks of behavior are lacking.




\section{Dataset}

Human decision-making under risk and uncertainty is often studied through tasks involving choices between \(m\) lotteries (or gambles), \(\{L_{1}, L_{2}, \ldots, L_{m}\}\), where for each \(i \in [m]\), \(L_{i}\) is defined by \(N\) possible payoffs \(\{x_i^m\}_{i=1}^N\) and their respective probabilities \(\{p_i^m\}_{i=1}^N\). Whereas traditionally, the options' payoff distributions are explicitly and numerically described, we study choices where these lotteries are described using free text. The dataset we use, \textbf{\textdataset{}} \cite{erevinprep}, includes 1,000 tasks of choice between $m=2$ lotteries labeled \textit{Option A} and \textit{Option B}. To create this dataset, (numeric) payoff distributions for the choice tasks were first randomly sampled from a large space. Then, an LLM converted these distributions to natural language, avoiding direct references to specific payoffs or probabilities.  Multiple descriptions were generated for each option, with one randomly selected for inclusion in the dataset. Further details on the creation of the dataset is given in \cite{erevinprep} and in Part \ref{dataset_sm} of the online \textcolor{blue}{\href{https://osf.io/95raf/?view_only=eb9c5a67baa4411f8d1b09544ae8f95d}{Supplementary Material}} (SM). 

Each textually-described choice task was completed by, on average, 31 participants, recruited using Prolific. Each participant completed 5 tasks, making a single decision without feedback. Participants were incentivized to maximize earnings: Their bonus payment depended on the realized payoffs from the options they selected. Our study aims to predict the proportion of participants who chose \textit{Option A} based solely on the textual descriptions of each option. 

\section{Method}
\label{sec:method}

We explored four complementary approaches to using large language models (LLMs) to predict human decisions under risk: (1) fine-tuning LLMs \cite{binz2024centaur, jeong2024fine} on \textbf{\textdataset{}}; (2) leveraging pre-trained embeddings with regression models \cite{binz2023turninglargelanguagemodels}; (3) "LLMs as subjects" - prompting LLMs to act as decision makers \cite{liu2024largelanguagemodelsassume, shapira2024can}; and (4) extracting interpretable behavioral features that past research has suggested are central to choice under risk and uncertainty from the textural descriptions. 

\subsection{Experimental Setup}
We used \textbf{\textdataset{}} as the main dataset throughout our experiments. Across all approaches, we allocated 90\% of the dataset (\(N=900\)) for training and validation, reserving 10\% (\(N=100\)) as a fixed held-out test set used consistently for model evaluation. Model selection (e.g., hyperparameter tuning, early stopping) was conducted using the validation subset of the training set. 
For all predictive models, we report: (a) mean-squared error (MSE) between predicted and observed proportions of Option A choices, and (b) directional accuracy—i.e., whether the prediction and observed label fall on the same side of the 0.5 threshold.

Some of our approaches involved training regression models to predict human choices based on either data representations (e.g., embeddings) or structured outputs from LLMs. We evaluated a range of regression techniques, including Linear Regression, Ridge, Lasso, Support Vector Regression (SVR), XGBoost, K-Nearest Neighbors (KNN), and Multi-Layer Perceptrons (MLPs). While all were explored during development, only the most relevant and best-performing models are reported in the results section.

\subsection{Incorporating Psychological Theory}
\label{subsec:psych-theory}

Recent research highlights the benefits of hybrid models that combine psychological theory with machine learning techniques to improve both predictive accuracy and interpretability. In this work, we investigate several such integrations, focusing on the BEAST model (Best Estimate and Sampling Tools) \cite{erev2017anomalies}—a highly successful behavioral model developed to explain and predict decisions under risk and uncertainty. BEAST models choice as the outcome of a partially biased mental sampling process, modulated by sensitivity to expected values. It has been shown to capture 14 well-documented choice anomalies and has won two international choice prediction competitions \cite{plonsky2019predicting, erev2017anomalies}. 

Furthermore, BEAST has served as the foundation for previous hybrid approaches that combined psychological features with machine learning algorithms to predict human choice in numerically described lotteries \cite{plonsky2017psychological, bourgin2019cognitivemodelpriorspredicting, plonsky2019predicting}. Following this line of work—which has achieved state-of-the-art results on the largest available datasets—we adopt BEAST as the core theoretical framework for guiding and augmenting our prediction models. In what follows, we describe how BEAST is incorporated into multiple components of our workflow. 
We incorporated BEAST in three ways:

\begin{enumerate}
    \item \textbf{BEAST-labeled pretraining:} In addition to fine-tuning solely on \textbf{\textdataset{}} (Approach 1), we also experimented with pretraining models on a synthetic dataset (N=20{,}000) of numerically described choices labeled by BEAST.
    \item \textbf{Personality-driven prompting:} For LLMs-as-subjects (Approach 3), we designed prompts to encode BEAST-inspired “personalities” aligned with cognitive dimensions.
    
    \item \textbf{Feature extraction:} In Approach 4, we proposed an alternative approach that used LLMs to extract BEAST-inspired features from tasks, following \cite{plonsky2017psychological}, and trained a regression model on these representations.
\end{enumerate}


\subsection{Approach 1: Fine-Tuned LLMs}
We fine-tuned multiple pre-trained LLMs based on the training data. We utilized BERT-based models, including BERT \cite{devlin2019bert}, RoBERTa \cite{liu2019roberta}, and DeBERTa \cite{he2021deberta}, due to their ability to generate rich, context-aware text representations that are well-suited for regression and predictive modeling. Additionally, we trained OpenAI's GPT-4o and GPT-4o-mini \cite{openai2023gpt4}, leveraging their advanced capacity to interpret complex textual patterns and perform qualitative reasoning, making them highly adaptable across diverse predictive scenarios.

Some fine-tuned models, such as GPT-4o and GPT-4o-mini, produce stochastic outputs during inference. To mitigate this variability, we generated 10 predictions for each sample and averaged them. Formally, for a sample \(i\), the prediction in these cases is: $\hat{p}_{i} = \frac{1}{10} \sum_{j=1}^{10} \hat{p}^{j}_{i}$. 

\subsubsection{Additional Training Data}
Because the size of the \textbf{\textdataset{}} dataset is limited, we explored two strategies for incorporating additional data into the training (i.e. fine-tuning) phase. Notably, to our knowledge, no other dataset of choice between textually described options exists. We thus chose to supplement the pre-training phase with data on choice between numerically described lotteries. First, we pre-trained our model with real data on choices between lotteries, using the large numerical dataset, \textbf{\petersondataset{}} \cite{peterson2021using}. Here, we used the $1039$ choice tasks that do not include feedback and ambiguity, to align with our experimental setting. Of these, we used 935 for training and validation and $104$ as the held-out test set. Second, we also tried pretraining using a large synthetic dataset that we generated specifically for this study (N = 20,000). The labels for this dataset were derived from the BEAST model \cite{erev2017anomalies}, a strong behavioral model rooted in psychological theory.

\subsection{Approach 2: Text Embeddings}
We transformed the textual data into numerical embeddings using OpenAI's \textit{text-embedding-ada-002} and \textit{text-embedding-3-large} models. These embeddings capture semantic relationships in continuous vector spaces, enabling downstream regression tasks. \textit{text-embedding-ada-002} emphasizes efficiency and cost-effectiveness, while \textit{text-embedding-3-large} offers richer semantic representation with higher dimensionality. For each task, we transformed the description of each option into an embedding vector, denoted as \(\mathbf{v}_A\) for \textit{Option A} and \(\mathbf{v}_B\) for \textit{Option B}. To capture the relationship between the options, we computed the task representation as the difference between the two embedding vectors: $\mathbf{d} = \mathbf{v}_A - \mathbf{v}_B$ where \(\mathbf{d}\) represents the embedding difference vector for the task. Using this task representation,\footnote{Other representations were tested, but we focused on vector difference as it performed best.} we applied various regression techniques, as described above, to predict the outcome.

We also investigated the effect of using PCA to reduce the dimensionality of the embedded vectors on regression performance. Dimensionality reduction helps mitigate computational costs and overfitting, especially with high-dimensional data. PCA transforms data into a set of orthogonal components ranked by their contribution to variance. Using PCA, we retained 5\%, 10\%, 25\%, and 33\% of the original dimensions and evaluated the trade-off between model complexity and predictive accuracy across these dimensions.

\subsection{Approach 3: LLMs as Simulated Subjects}
We designed an experimental framework where LLM agents acted as \textit{``experimental subjects"}. Each agent faced and provided its choices for 50 of the choice tasks (See Figures ~
~\ref{fig:binary_prompt}, ~\ref{fig:confidence_prompt}, ~\ref{fig:percentage_prompt}, in the \textcolor{blue}{\href{https://osf.io/95raf/?view_only=eb9c5a67baa4411f8d1b09544ae8f95d}{SM}}. The responses from all agents were aggregated for each task to generate the final LLM's prediction. Then, a regression model was trained to learn the relationship between the LLM's predictions and human choices, providing an optimized mapping between the two. 

\subsubsection{Prompting Conditions}

The LLM agents' responses were elicited under three distinct prompting conditions. In the {\textit{Binary}} condition, the LLM made a direct choice between the two options. In the {\textit{Percentage}} condition, the LLM provided a continuous preference score between 0 and 100. Finally, the {\textit{Confidence}} condition required the LLM to make a binary choice and then assign that choice a confidence level (0–100), which was used for predictions.

\subsubsection{Personalities}

To investigate the influence of psychological theory on the model’s performance, we developed ten distinct \textit{Personalities}, each reflecting an assumption (or a combination of assumptions) embedded in the BEAST model \cite{erev2017anomalies}. For instance, one of BEAST's assumptions is that people are sometimes more sensitive to the sign of the reward (gain or loss) than the actual values. Accordingly, one of the personalities is \textit{The Guardian}, which was defined to behave as someone who is "Sensitive to gains vs. losses, impacting risk tolerance". The interpretations and details of these personalities are presented in Table ~\ref{table:beast_persona} in the \textcolor{blue}{\href{https://osf.io/95raf/?view_only=eb9c5a67baa4411f8d1b09544ae8f95d}{SM}}. 

For comparison, we included a baseline model where all agents operated without any assigned personality. To improve predictions, we aggregated the outputs from each predefined personality profile and trained a weighted regression model, where each personality contributes to the final prediction according to its optimized weight. This approach captures the collective predictive power of the different personalities while accounting for their unique contributions to overall prediction performance.

\subsection{Approach 4: Theory-Guided Feature Extraction}
We used the LLM to extract from the textual descriptions numeric values for behavioral features, transforming the task into a numerical prediction task with a well-established solution. Building on the work of \citeA{plonsky2019predicting}, which demonstrated that human choices can be effectively predicted using ML and features derived from the behavioral model BEAST, we aimed to extract a set of features that capture various elements of BEAST. For instance, one of BEAST’s assumptions is that people tend to exhibit \textit{pessimism}, expecting the worst possible outcome. To reflect this, we extracted a "worst-case" feature, which identifies the option with the better payoff under the worst-case scenario. All the extracted features appear in  Table~\ref{tab:decision_prompts} in the \textcolor{blue}{\href{https://osf.io/95raf/?view_only=eb9c5a67baa4411f8d1b09544ae8f95d}{SM}}.

The primary objective was not to assess the accuracy of the LLM's feature extraction but to ensure that its process mirrored human-like reasoning. For instance, when a description emphasized disadvantages, it was reasoned that human subjects might ``extract" a set of perceived values different from the actual numerical values (which were unknown to them) and base their decisions on these perceptions.

To implement this, we designed specific prompts for each feature and instructed the LLM to classify which option was preferred under the assumption of that feature. To account for ambiguity, we allowed the LLM to provide a neutral response when no clear preference could be inferred. The results were aggregated and converted into numeric scores, which were then trained using a regression model (as mentioned above) for final predictions.

\subsection{Evaluation on Numerical Tasks}

We also evaluated how some of the best models perform with tasks involving numeric descriptions. To do so, we used the \textbf{\petersondataset{}} \cite{peterson2021using} dataset, the largest dataset of risky choice publicly available. Of this dataset, we used the subset of tasks that excluded feedback and ambiguity to match our experimental conditions. 90\% (N = 935) of this set was used for training and validation while the rest of the data (N=104) was used as a held-out set. We fine-tuned GPT-4o on this dataset and, as a benchmark, also trained BEAST-GB \cite{plonsky2019predicting}, which is currently considered state-of-the-art in this numerical description setting.

\section{Results}
\begin{table*}[t]
\centering
\caption{Models Performance on \textbf{\textdataset{}} (Textually Described Choices)}
\label{tab:main_res}
\vskip 0.12in
\small
\setlength{\tabcolsep}{6pt}
\renewcommand{\arraystretch}{1.2}
\begin{tabular}{lllcc}
\toprule
\textbf{Approach} & \textbf{Model} & \textbf{Training Data} & \textbf{Test MSE} & \textbf{Test Accuracy} \\
\midrule

\multirow{7}{*}{\textbf{Fine-Tuning}} 
  & BERT                & Textual only                        & 0.0260 & 0.84 \\
  & RoBERTa             & Textual only                        & 0.0169 & 0.87 \\
  & DeBERTa             & Textual only                        & 0.0162 & 0.92 \\
  & GPT-4o-mini\textsuperscript{*} & Textual only          & 0.0130 & 0.87 \\
  \cmidrule(lr){2-5}
  & \multirow{3}{*}{GPT-4o\textsuperscript{*}} 
                        & Textual only                        & 0.0121 & 0.87 \\
  &                    & Textual + Numerical                 & \textbf{0.0110} & 0.88 \\
  &                    & Textual + Synthetic BEAST           & 0.0123 & 0.85 \\

\midrule
\multirow{2}{*}{\textbf{Embeddings}} 
  & MLP
    & Textual only                        & 0.0138 & 0.89 \\
  & Ridge
    & Textual only                        & 0.0159 & 0.88 \\

\midrule
\multirow{2}{*}{\textbf{LLM as Subjects}} 
  & Out-of-box LLM
    & --                                  & 0.0170 & 0.87 \\
  & BEAST-personalities LLM
    & --                                  & 0.0220 & 0.81 \\

\midrule
\textbf{Feature Extraction} 
  & XGBRegressor
    & BEAST-derived model                 & 0.0395 & 0.73 \\

\bottomrule
\end{tabular}

\vspace{0.15cm}
\parbox{0.95\textwidth}{%
\raggedright
{\footnotesize \textit{Note:} \textsuperscript{*} Results are based on stochastic models averaged over 10 inference runs.}
}
\end{table*}

\begin{table}[ht]
\centering
\caption{MSE by target extremity.}

\label{tab:error_analysis}
\vskip 0.12in
\small
\setlength{\tabcolsep}{6pt}
\begin{tabular}{lcc}
\toprule
\textbf{Model} & \textbf{Extreme (n=29)} & \textbf{Non-Extreme (n=71)} \\
\midrule
GPT-4o    & 0.0066 & 0.0128 \\
Embedding & 0.0136 & 0.0127 \\
DeBERTa   & 0.0140 & 0.0172 \\
\bottomrule
\end{tabular}
\parbox{0.95\textwidth}{
{\footnotesize \textit{Note:} \textit{Extreme} denotes tasks with target values below 0.2 \\ or above 0.8.}}
\end{table}

\begin{table}[ht]
\centering
\caption{\centering Comparison of model's performance on Numeric Dataset \textbf{\petersondataset{}}}
\vskip 0.12in

\label{tab:numeric_results}
\setlength{\tabcolsep}{4pt} 
\renewcommand{\arraystretch}{1.2} 
\small 
\begin{tabular}{lcc}
\hline
\rowcolor[HTML]{EFEFEF} 
\textbf{Model} & \textbf{Test MSE} & \textbf{Test Accuracy}\\ 
\hline 
BEAST-GB           & 0.0094 & 0.89 \\ 
GPT-4o      & 0.0104   & 0.89 \\                                   
\hline
\end{tabular}
\end{table}

We present the results of all different models we tried in the \textcolor{blue}{\href{https://osf.io/95raf/?view_only=eb9c5a67baa4411f8d1b09544ae8f95d}{SM}}.  Here, we focus on the best set of models within each approach. Table \ref{tab:main_res} presents these results.\footnote{A previous version of this paper mistakenly reported different results due to a data processing error.} Fine-tuning of LLMs outperformed alternative methods, highlighting its effectiveness in adapting pre-trained linguistic representations to the task. Fine-tuning a GPT-4o model achieved strong results, with an MSE of \texttt{0.0121} and an accuracy of \texttt{0.87}. Incorporating numerical data into the training process further improved performance, reducing the MSE to \texttt{0.0110}, the best result achieved in terms of MSE. In contrast, adding synthetic BEAST data slightly degraded performance, increasing the MSE to \texttt{0.0123}.

Using embeddings extracted from textual descriptions and training traditional machine learning models such as MLP and Ridge regression represented the best non-fine-tuning approach, achieving MSEs of \texttt{0.0138} and \texttt{0.0159}, respectively. The ``LLM as Subjects" approach, where out-of-the-box LLMs or BEAST-personalities were prompted directly, resulted in higher MSEs (\texttt{0.0170} and \texttt{0.0220}). Feature extraction based on BEAST-derived representations performed worst, with a relatively high MSE of \texttt{0.0395}.


Other fine-tuned models are also presented in Table \ref{tab:main_res}. GPT-4o-mini, the smaller variant of GPT-4o, performed slightly worse than the larger model, achieving an MSE of \texttt{0.0130} with a similar accuracy of \texttt{0.87}, yet still outperforming all other models. Traditional transformer models such as RoBERTa and DeBERTa demonstrated decent performance, with RoBERTa reaching an MSE of \texttt{0.0169} and DeBERTa achieving \texttt{0.0162}. BERT lagged behind, with a substantially higher MSE of \texttt{0.0260}. Interestingly, the accuracy of DeBERTa was found to be highest of all models we tried.

We further analyzed the errors of three models: GPT-4o, our best overall model; the best embeddings model, which was the second-best approach; and DeBERTa, chosen for its high directional accuracy. Table~\ref{tab:error_analysis} shows that for tasks with non-extreme target values (i.e., between 0.2 and 0.8), GPT-4o and the embeddings model performed similarly but in tasks with extreme target values, GPT-4o was clearly better. DeBERTa also performed better on extreme tasks than on non-extreme ones. These results suggest that GPT-4o's overall advantage may stem from its ability to handle extreme decisions more effectively.

Finally, to evaluate the robustness of fine-tuned LLMs when applied to numerically described gambles, we tested the best models on the numeric dataset \petersondataset{} (Table \ref{tab:numeric_results}). Here, GPT-4o maintained relatively strong performance with an MSE of \texttt{0.0104} and an accuracy of \texttt{0.89}. However, it still underperformed compared to BEAST-GB \cite{plonsky2019predicting}, a hybrid model combining behavioral theories with ML, which achieved a lower MSE of \texttt{0.0092}.

\section{Discussion}

Human choice under risk has been extensively studied for decades, but this research has predominantly focused studying tasks with accurate numeric descriptions. This approach, while valuable, does not fully capture the richness and complexity of real-world decisions, which often involve potentially ambiguous textual information. We take an important step by examining choice behavior in textually described contexts, offering a closer approximation of how people navigate decisions in naturalistic settings. Our findings reveal important differences between these two domains, highlighting their distinct challenges and opportunities for behavioral theories and for ML models.

Our findings reveal a significant gap between textual and numeric decision-making tasks. While theory-free ML approaches excelled in the textual domain, a hybrid of behavioral theories and ML, specifically BEAST-GB,  demonstrated its continued advantage in the numeric setting. This discrepancy highlights potentially fundamental differences in how textual and numeric data are processed. Textual descriptions often include interpretive ambiguity, allowing language models to leverage fine-tuning for task-specific optimization. Numeric data, by contrast, benefits from the structured assumptions provided by behavioral theories, which align well with predefined, explicit representations of choices.

We find that fine-tuning GPT-4o achieved the best performance on \textbf{\textdataset{}}. These results became even stronger when we incorporated additional pretraining on numerical data, which may imply that despite the aforementioned potential differences between the underlying processes involved in choices between textually and numerically described lotteries, they also share some similarities, and people's choices in one type of tasks are associated with their choices in the other. Future research should focus on the key similarities and differences between the two types of tasks. 

Furthermore, GPT-4o demonstrated remarkable robustness across both textual and numerical tasks. Although it did not achieve the top score in the numerical domain, where BEAST-GB outperformed it, it still performed competitively. This resilience, particularly when leveraging numerical and BEAST synthetic data, highlights GPT-4o’s ability to handle diverse and noisy data sources. We attribute this strength to its broader pretraining and superior generalization capacity. Overall, these findings underscore GPT-4o’s versatility and position it as a strong candidate for general-purpose decision-making tasks across a wide range of domains.

Despite the success of BEAST-GB in numeric tasks, attempts to integrate psychological theory into textual decision-making were less effective. BEAST-derived models and synthetic data did not enhance performance compared to theory-free versions of the same models. Feature extraction, which is a fully theory-driven method approach performed particularly poorly. This is surprising given that psychological theory has historically improved predictive accuracy in numeric settings. One possible explanation is that the richness and complexity of textual data dilute the utility of predefined behavioral constructs, which are inherently designed for structured numeric inputs. It is important to note that all our approaches to incorporate psychological theory were based on the model BEAST. Hence, our results do not necessarily imply that integrating behavioral theory based on different models or theories would also be ineffective. However, BEAST has a strong track record in numerical settings and, even in our analysis, BEAST-based models outperformed all other models, highlighting its strengths in structured, quantitative tasks. Furthermore, when using BEAST as a foundation for LLM personalities or feature extraction, our approach may not have effectively captured key elements of the model, as some aspects are non-trivial to process. Adapting such frameworks to qualitative, language-based representations remains a significant challenge.

These results underscore the need to develop hybrid models better suited for textual tasks, combining insights from behavioral theories with the capabilities of modern LLMs. One promising avenue is to refine feature engineering to align behavioral constructs with the nuances of textual data. Additionally, exploring how LLMs process qualitative, ambiguous information could yield valuable insights into computational decision-making models. Future research should also investigate how task-specific fine-tuning can be further optimized to bridge the gap between textual and numeric settings.

While this study provides valuable insights, some limitations should be noted. The relatively small size of \textbf{\textdataset{}} may limit the generalizability of the findings, particularly for complex models like GPT-4o. Additionally, the inherent differences between controlled numeric tasks and naturalistic textual descriptions may pose challenges for direct comparisons. Finally, it is important to acknowledge that we have not tested all possible LLMs, and as this field evolves rapidly, more advanced models may already exist or emerge in the near future. This highlights the need for ongoing research to evaluate and compare the latest advancements in ML for decision-making tasks.

\subsection{Conclusion}

Our work highlights the effectiveness of task-specific fine-tuning for textual decision-making tasks, with GPT-4o achieving state-of-the-art performance. However, the gap between textual and numeric settings, along with the challenges of incorporating psychological theory, points to the need for further research. By bridging these gaps, future studies can advance our understanding of human decision-making and improve the predictive capabilities of computational models.

\bibliographystyle{apacite}
\setlength{\bibleftmargin}{.125in}
\setlength{\bibindent}{-\bibleftmargin}
\bibliography{Bib}

\newpage
\onecolumn
\appendix

\renewcommand{\thetable}{A.\arabic{table}}
\renewcommand{\thefigure}{A.\arabic{figure}}
\setcounter{table}{0}
\setcounter{figure}{0}
\noindent\textbf{{\Huge Supplementary Material}}
\part{Dataset Information} \label{dataset_sm}

We employed a large language model (LLM) using three distinct methods to convert numerical gamble descriptions into natural language. Each method focused on a different strategy for generating textual descriptions, leveraging both theoretical principles and task-specific numerical characteristics. For each option in each decision-making task, we generated multiple verbal descriptions across all methods. For the final dataset, one description per option was randomly selected.

\section{Method 1: Theory-Based Textual Labels}

We deterministically labeled each option in every task using simple descriptions derived from psychological theory (e.g., \textit{"Gamble can yield a positive payoff"}, \textit{"Gamble is better than the other gamble most of the time"}, etc.). Table \ref{tab:simple_descriptions} provides the complete list of these descriptions. We then instructed the LLM to generate a description for each unique set of labels and matched these descriptions to all tasks that shared the same labels. Next, to tailor the descriptions more closely to specific tasks, we added a modification stage where the language model adjusted the description, if necessary, based on the task's numerical values. This was done in two versions: one considering the values as they are, and the other focusing on the relationship between the options in the task (e.g., stating that, with probability \( p \), gamble A gives \( x \) more than B). All prompts are included in \ref{tab:generation_prompt}.

\begin{longtable}[h!]
{|p{0.95\textwidth}|}
\hline
\textbf{Method 1: Initial Generation} \\
\hline
I have a large dataset of problems, each row representing a pair of gambles in the following form: \\

Gamble A: a1 w.p. pa1 else a2 \\
Gamble B: b1 w.p. pb1 else b2 \\

Each gamble in the dataset is associated with a set of descriptive labels. I need help creating descriptions for these gambles based on their labels. Each label set should be transformed into a coherent and engaging sentence that conveys the essence of the gamble, with the context of all labels integrated implicitly or explicitly. The descriptions should:

\begin{itemize}[label=-]
    \item Be clear and accessible to someone without expert knowledge in the field.
    \item Be free of technical jargon, aiming for an engaging and informative style.
    \item Be varied in language and structure, avoiding repetition.
    \item Provide a holistic understanding of each gamble, reflecting the broader context provided by all the labels.
    \item Each description should integrate all or some of these labels into a single sentence, using a maximum of 30 words. The goal is to provide a deeper, more nuanced understanding of each gamble through these descriptions.
    \item Some descriptions may be negative and include disadvantages and reasons that might make someone choose against it.
\end{itemize}

Please generate 10 different descriptions for each set of labels, with each description being just one sentence. \\

The label sets are: \\

[Sets of simple labels] \\

Return results as JSON without any extra text. \\
\hline
\textbf{Method 1: Absolute Values Modification} \\
\hline
\textbf{Objective}: Systematically evaluate and, where necessary, revise a list of descriptions for gambles within Task [Task ID]. Each description pertains exclusively to either Gamble A or Gamble B. Revisions should be made only if a description inaccurately represents its respective gamble's nature or scale. The revised descriptions must be realistic, contextually fitting, avoid specific numbers, and should not directly mention "Gamble A" or "Gamble B". However, each should subtly reflect the gamble's comparative advantages or disadvantages against the alternative, based on the provided data. If a description is accurate as is, it should be noted as "No Modification".
 \\

\textbf{Context}: The original descriptions, generated from labels without access to detailed values, may not fully capture the gambles' outcomes. They often highlight relative strengths or weaknesses, a feature to be preserved if supported by the data. \\

\textbf{Data Provided:
} \\

\ \ 1. Original Description for Gamble [A/B] of Task [Task ID]: [original description] \\ 
\ \ 2. Gamble Values: \\
\ \ \ \ \ \ For Gamble A: Outcome of [a1] with   probability [pa1], else [a2] \\
\ \ \ \ \ \ For Gamble B: Outcome of [b1] with probability [pb1], else [b2] \\
\ \ 3. Labels Used for Original Description: [List the labels that were originally used] \\

\textbf{Expected Output}: A list of revised descriptions, without original descriptions. For each description in the list, provide a subtly adjusted description of around 20 words that accurately conveys the provided values for its respective gamble. The revision should imply its comparative position relative to the alternative gamble without using explicit labels like "Gamble A" or "Gamble B". Avoid numerical specifics, ensuring the description remains grounded in realism and context. For descriptions that are found to be accurate, return "No Modification". \\

\hline
\textbf{Method 1: Relative Values Modification\footnote{Note that when using the prompt, not all combinations regarding the relationship between gambles, as shown in this example, were displayed; only the options that were possible based on the values of the two options were shown.}} \\
\hline
\textbf{Objective}: Systematically evaluate and, where necessary, revise a list of descriptions for gambles within Task [Task ID]. Each description pertains exclusively to either Gamble A or Gamble B. Revisions should be made only if a description inaccurately represents its respective gamble's nature or scale. The revised descriptions must be realistic, contextually fitting, avoid specific numbers, and should not directly mention "Gamble A" or "Gamble B". However, each should subtly reflect the gamble's comparative advantages or disadvantages against the alternative, based on the provided data. If a description is accurate as is, it should be noted as "No Modification". \\

\textbf{Context}: The original descriptions, generated from labels without access to detailed values, may not fully capture the gambles' outcomes. They often highlight relative strengths or weaknesses, a feature to be preserved if supported by the data. \\

\textbf{Data Provided: 
} \\

1. Original Description for Gamble [A/B] of Task [Task ID]: [original descriptions] \\ 
2. Gamble [A/B] gives:  \\
\ [x1] less than [B/A] with probability [p1] \\ 
\ [x2] less than [B/A] with probability [p2] 
\\
\ [x3] more than [B/A] with probability [p3] 
\\
\ [x4] more than [B/A] with probability [p4] 
\\
\ The same outcome as [B/A] with probability [p5] 
\\ 
3. Labels Used for Original Description:  [List the labels that were originally used] \\ 

\textbf{Expected Output}: A list of revised descriptions, without original descriptions. For each description in the list, provide a subtly adjusted description of around 20 words that accurately conveys the provided values for its respective gamble. The revision should imply its comparative position relative to the alternative gamble without using explicit labels like "Gamble A" or "Gamble B". Avoid numerical specifics, ensuring the description remains grounded in realism and context. For descriptions that are found to be accurate, return "No Modification". \\

\hline
\textbf{Method 2} \\
\hline
Extract the main clause of each sentence: [list of sentences] and return results as a list. \\

\hline
\textbf{Method 3} \\
\hline
There is a pair of gambles in the following format:

Gamble A: [A]
Gamble B: [B]

Your job is to provide verbal descriptions of the gambles. State clearly which description is for which gamble. These descriptions will later be presented to people who will need to choose between these gambles. \\ 

I want for each gamble separately, [k] distinct descriptions from each of the following four categories: \\
\\ 
1. Neutral: description will be balanced, stating both advantages and disadvantages of the gamble. We will use such description when we do not want to influence people for either choosing or not choosing the gamble. \\
2. Advantage Focused: description should focus on the gamble's advantages (perhaps in relation to the alternative - see below). We will use such description when we want to make people more likely to choose the gamble. \\
3. Disadvantage Focused: description should focus on the gamble's disadvantages (perhaps in relation to the alternative - see below). We will use such description when we want to make people less likely to choose the gamble. \\
4. Is Better: for each gamble, description should reflect which gamble is "better" and in what way is "better". Note that sometimes one gamble is always better than the other (has dominance over the other) please address that. \\

\textbf{The Rules:
} \\
1. Descriptions must accurately convey the nature of each gamble, including its potential outcomes and probabilities \\
2. You must not directly lie, but you may give only a partial image of the gamble, e.g., by not stating its disadvantages or only emphasizing its advantages \\
3. Use only words and avoid using numbers. \\
4. You can and often should provide a description that takes into account the relationship between the two gambles, e.g., by referring to a gamble's advantages or disadvantages in comparison with its alternative. \\
5. Avoid using the phrases "Gamble A" or "Gamble B" in your description, or any other naming convention. Simply give the description itself. \\
6. Sometimes one gamble has dominance over the other, if it is true you may address that in your description. \\

Generate a structured response in JSON format, in the following structure:

\{ \\
\ \ "Gamble A": \{ \\
\ \ \ \ "Neutral": ["string", "string", ...], \\
\ \ \ \ "Advantage Focused": ["string", "string", ...], \\
\ \ \ \ "Disadvantage Focused": ["string", "string", ...], \\
\ \ \ \ "Is Better": ["string", "string", ...] \\
\ \ \}, \\
\ \ "Gamble B": \{ ... \} \\
\} \\

\hline
\caption{Prompts Data Generation}
 \label{tab:generation_prompt} 

\end{longtable}
\begin{table}[h!]
\centering

\caption{List of Simple Descriptions for Method 1} \label{tab:simple_descriptions} 
\begin{tabularx}{\textwidth}{|>{\raggedright\arraybackslash}X|}
\hline
\textbf{Label} \\
\hline
Gamble has higher variance compared to the other gamble \\
Gamble has lower variance compared to the other gamble \\
Gamble is never worse than the other gamble \\
Gamble is never better than the other gamble \\
Gamble is better than the other gamble most of the time \\
Gamble has a higher payoff than the other gamble on average \\
Gamble has a lower payoff than the other gamble on average \\
Gamble cannot yield a negative payoff \\
Gamble cannot yield a positive payoff \\
Gamble always yields a payoff of zero \\
Gamble can yield a positive payoff \\
Gamble has a positive average payoff \\
Gamble holds the maximum value among all possible outcomes \\
Gamble holds the minimum value among all possible outcomes \\
Gamble has a higher probability of a positive payoff compared to the other gamble \\
Gamble has a lower probability of a positive payoff compared to the other gamble \\
Gamble can yield a negative payoff \\
Gamble has a negative average payoff \\
Gamble has a higher probability of a negative payoff compared to the other gamble \\
Gamble has a lower probability of a negative payoff compared to the other gamble \\
Gamble yields constant positive payoff \\
Gamble always yields a positive payoff \\
Gamble has a high probability of yielding a positive payoff \\
Gamble yields constant negative payoff \\
Gamble always yields a negative payoff \\
Gamble can lead to a large positive payoff \\
Gamble has a high probability of yielding a negative payoff \\
Gamble has a low probability of yielding a positive payoff \\
One of the outcomes of the gamble has a low probability \\
Gamble is always better than the other gamble \\
Gamble is always worse than the other gamble \\
Gamble can lead to a large negative payoff \\
Gamble has a low probability of yielding a negative payoff \\
\hline
\end{tabularx}
\end{table}

\section{Method 2: Main Clause Extraction}

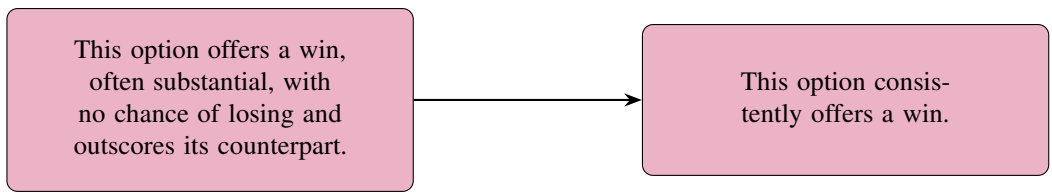
\begin{figure}[h!]
\centering
\begin{tikzpicture}[
  box/.style={draw, rounded corners, fill=purple!30, text width=4.5cm, align=center, minimum height=2cm, inner sep=12pt}, 
  arrow/.style={->, thick, >=Stealth},
  node distance=2cm and 2cm 
]

\node[box] (left) {This option offers a win, often substantial, with no chance of losing and outscores its counterpart.};
\node[box, right=3cm of left] (right) {This option consistently offers a win.};

\draw[arrow] (left.east) -- ++(0.5,0) -- ++(1,0) |- (right.west);

\end{tikzpicture}
\caption{Shortening Descriptions By Extracting the Main Clause}
\label{fig:shorten_modification}
\end{figure}

Since many of the descriptions generated in \textit{Method 1} were very long, we decided to introduce another method where we took the descriptions generated in \textit{Method 1} and instructed the LLM to extract the main clause of each description. Figure \ref{fig:shorten_modification} shows an example of this process.

\section{Method 3: Category-Based Description Generation}

Based on the numeric values of each task, we instructed the LLM to generate descriptions for four different categories:
\newline 

\begin{tabularx}{\textwidth}{lX}
\textbf{Category} & \textbf{Description} \\
\hline
\textit{Neutral} & Balanced descriptions that do not attempt to influence selection. \\
\textit{Advantage Focused} & Descriptions that encourage the subject to choose the gamble. \\
\textit{Disadvantage Focused} & Descriptions that discourage the subject from choosing the gamble. \\
\textit{Is Better} & Descriptions that indicate which gamble is "better." \\
\end{tabularx}

\part{Predictive Models}

\section{Fine Tuning LLM}

\subsection{Prompt}

\begin{figure}[H]
    \centering
    \fbox{
        \begin{minipage}{0.9\textwidth}
            Estimate the percentage of the population choosing Option A over Option B: \\
            \textbf{Option A:} \{A\} \\
            \textbf{Option B:} \{B\}
        \end{minipage}
    }
    \caption{Prompt used for Fine-Tuning LLM}
    \label{fig:llm_tuning_prompt}
\end{figure}

\subsection{Training Details}
All Open-AI's models were fine-tuned using OpenAI's fine-tuning API via the OpenAI platform (website interface). 
All models were trained for \textbf{5 epochs}. 
The training details are summarized in Table~\ref{table:gpt4o_tuning}. 
 Fine-tuned model references are available upon request.

All BERT-family models were fine-tuned using the \textbf{Hugging Face Trainer} framework on \textbf{Google Colab computing units}. A detailed breakdown of the hyperparameters used for each model is presented in Table~\ref{table:bert_family_tuning}.

\begin{table}[ht]
\centering
\caption{GPT-4o and GPT-4o-mini Tuning Details}
\label{table:gpt4o_tuning}
\begin{tabular}{|l|c|c|}
\hline
\rowcolor[HTML]{EFEFEF} \textbf{Training Data}  & \textbf{Batch Size} & \textbf{lr Multiplier}  \\
\hline
\multicolumn{3}{|c|}{\textbf{GPT-4o}} \\ 
\hline
Verbal         & 2 & 2  \\
Numeric        & 2 & 2     \\
Numeric-Verbal & 2 & 2     \\
BEAST          & 2 & 66    \\
BEAST-Verbal   & 2 & 2     \\
BEAST-Numeric  & 2 & 2    \\
\hline
\multicolumn{3}{|c|}{\textbf{GPT-4o-mini}} \\ 
\hline
Verbal         & 4 & 1.8  \\
\hline
\end{tabular}
\end{table}

\begin{table}[ht]
\centering
\caption{BERT Family Fine-Tuning Details}
\label{table:bert_family_tuning}
\begin{tabular}{|l|c|c|c|c|c|}
\hline
\rowcolor[HTML]{EFEFEF} \textbf{Model} & \textbf{Dataset} & \textbf{Epochs} & \textbf{Batch Size} & \textbf{LR} & \textbf{Max Steps} \\
\hline
BERT      & Verbal Only      & 30  & 16  & 1.9333e-05 & 1500  \\
RoBERTa   & Verbal Only      & 30  & 16  & 1.9333e-05 & 1500  \\
DeBERTa & Verbal Only      & 30  & 16   & 1.9333e-05 & 1500  \\
\hline
\end{tabular}
\end{table}

\newpage

\subsection{Results}

\begin{table}[ht]
\centering
\begin{tabular}{|l|c|}
\hline
\rowcolor[HTML]{EFEFEF} 
\textbf{Model}                & \textbf{Test MSE} \\ \hline
BERT                          & 0.0167            \\ \hline
RoBERTa                       & 0.0110             \\ \hline
DeRoBERTa                     & 0.0110             \\ \hline
Ensemble GPT-4o-mini*        & 0.0130               \\ \hline
Ensemble GPT-4o*             & 0.0121               \\ \hline
\end{tabular}
\caption{\centering  *averaged over 10 inference runs.}
\label{tab:model_test_mse}
\end{table}

\begin{table}[h!]
\centering
\begin{tabular}{|c|c|c|c|c|c|}
\hline
\textbf{Epochs} & \textbf{Batch} & \textbf{MSE} & \textbf{Mean} & \textbf{Std} & \textbf{CI} \\ \hline
5 & 2  & 0.0132 & 0.0146 & 0.0009 & [0.01318 0.01601]     \\ \hline
5 & 4  & 0.0130 & 0.0144 & 0.0007 & [0.0135225 0.015555]  \\ \hline
5 & 8  & 0.0134 & 0.0161 & 0.0014 & [0.01418 0.0188175]   \\ \hline
5 & 16 & 0.0166 & 0.0203 & 0.0018 & [0.01718 0.0225875]   \\ \hline
\end{tabular}
\caption{Summary of Training Results: GPT-4o-mini. 10 Inference Runs.}
\label{tab:4o-mimi}
\end{table}

\begin{table}[h!]
\centering
\begin{tabular}{|c|c|c|c|c|c|}
\hline
\textbf{Epochs} & \textbf{Batch} & \textbf{MSE} & \textbf{Mean} & \textbf{Std} & \textbf{CI} \\ \hline
5 & 2  & 0.0121 & 0.014  & 0.0012 & [0.0124225 0.015455]    \\ \hline
5 & 4  & 0.0123 & 0.0146 & 0.0008 & [0.0135675 0.0158975]   \\ \hline
5 & 8  & 0.0136 & 0.0176 & 0.002  & [0.014205 0.0199325]    \\ \hline
\end{tabular}
\caption{Summary of Training Results: GPT-4o. 10 Inference Runs.}
\label{tab:4o}
\end{table}

\section{Text Embedding}

\subsection{Training Details}
\subsubsection{MLP}

To mitigate the effect of random seed variation, all reported results are averaged over \textbf{50 repetitions}. 
The best-performing model was trained on \textit{text-embedding-3-large} embeddings, with dimensionality reduced to approximately \textbf{33\%}. 
The mean error on the validation set was \textbf{0.0144}, with a \textbf{95\% confidence interval} of \textbf{[0.0138, 0.0150]}.  The selected hyperparameters are as follows:  
\begin{itemize}
    \item \textbf{Batch Size:} 64
    \item \textbf{Learning Rate:} 0.01
    \item \textbf{Weight Decay:} 0.001
    \item \textbf{Dropout Probability ($p$):} 0.6
    \item \textbf{Number of Layers:} 2
    \item \textbf{Hidden Dimensions:} [64, 128]
\end{itemize}

\subsubsection{Ridge Regressor}

The Ridge regression model was trained on \textit{text-embedding-3-large} embeddings, with dimensionality reduced to \textbf{153} (\textbf{5\%} of the original size). The validation mean squared error (MSE) was \textbf{0.0159}. The selected hyperparameters were \textbf{alpha = 1} and \textbf{fit\_intercept = True}.

\subsection{Results}

Tables \ref{tab:ada_embedding_results}, \ref{tab:large_embedding_results}
\ref{tab:ada_mlp} and \ref{tab:larg_mlp} present full results of all regression models that was fitted.

\begin{table}[H]
\centering
\caption{Model Performance Comparison Across Different Dimensions for Large Embedding}
\label{tab:larg_reg}
\begin{tabular}{|c|l|c|c|}
\hline
\textbf{Dimension} & \textbf{Model}              & \textbf{CV Score (MSE)} & \textbf{Test MSE} \\ \hline
\multirow{6}{*}{153 (~5\%)} & Linear Regression     & 0.0191                  & 0.0166           \\ 
                            & Ridge                & \textbf{0.0168}         & \textbf{0.0159}  \\ 
                            & Lasso                & 0.0171                  & 0.0161           \\ 
                            & SVR                  & 0.0181                  & 0.0176           \\ 
                            & XGBRegressor         & 0.0217                  & 0.0243           \\ 
                            & KNN                  & 0.0282                  & 0.0318           \\ \hline
\multirow{6}{*}{307 (~10\%)} & Linear Regression    & 0.0285                  & 0.0189           \\ 
                            & Ridge                & \textbf{0.0173}         & \textbf{0.0159}  \\ 
                            & Lasso                & 0.0178                  & 0.0160           \\ 
                            & SVR                  & 0.0184                  & 0.0172           \\ 
                            & XGBRegressor         & 0.0225                  & 0.0248           \\ 
                            & KNN                  & 0.0285                  & 0.0329           \\ \hline
\multirow{6}{*}{768 (~25\%)} & Linear Regression    & 0.1638                  & 0.0995           \\ 
                            & Ridge                & \textbf{0.0168}         & \textbf{0.0159}  \\ 
                            & Lasso                & 0.0180                  & 0.0162           \\ 
                            & SVR                  & 0.0183                  & 0.0173           \\ 
                            & XGBRegressor         & 0.0241                  & 0.0250           \\ 
                            & KNN                  & 0.0288                  & 0.0322           \\ \hline
\multirow{6}{*}{1000 (~33\%)} & Linear Regression   & 0.0286                  & 0.0333           \\ 
                            & Ridge                & \textbf{0.0168}         & \textbf{0.0160}  \\ 
                            & Lasso                & 0.0180                  & 0.0162           \\ 
                            & SVR                  & 0.0183                  & 0.0173           \\ 
                            & XGBRegressor         & 0.0240                  & 0.0265           \\ 
                            & KNN                  & 0.0288                  & 0.0323           \\ \hline
\multirow{6}{*}{3072 (~100\%)} & Linear Regression  & 0.0286                  & 0.0333           \\ 
                            & Ridge                & \textbf{0.0168}         & \textbf{0.0160}  \\ 
                            & Lasso                & 0.0180                  & 0.0191           \\ 
                            & SVR                  & 0.0183                  & 0.0173           \\ 
                            & XGBRegressor         & 0.0188                  & 0.0198           \\ 
                            & KNN                  & 0.0288                  & 0.0323           \\ \hline
\end{tabular}
\label{tab:large_embedding_results}
\end{table}

\begin{table}[H]
\centering
\caption{Model Performance Comparison Across Different Dimensions for Ada Embedding}
\label{tab:ada_reg}
\begin{tabular}{|c|l|c|c|}
\hline
\textbf{Dimension} & \textbf{Model}              & \textbf{CV Score (MSE)} & \textbf{Test MSE} \\ \hline
\multirow{6}{*}{76 (~5\%)}  & Linear Regression     & 0.0195                  & 0.0196           \\ 
                            & Ridge                & 0.019                   & 0.0209           \\ 
                            & Lasso                & 0.0187                  & 0.0198           \\ 
                            & SVR                  & 0.0252                  & 0.0255           \\ 
                            & XGBRegressor         & 0.0213                  & 0.0236           \\ 
                            & KNN                  & 0.027                   & 0.032            \\ \hline
\multirow{6}{*}{153 (~10\%)} & Linear Regression    & 0.0218                  & 0.018            \\ 
                            & Ridge                & 0.0188                  & 0.0198           \\ 
                            & Lasso                & 0.0189                  & 0.0183           \\ 
                            & SVR                  & 0.0226                  & 0.0201           \\ 
                            & XGBRegressor         & 0.0221                  & 0.025            \\ 
                            & KNN                  & 0.0276                  & 0.0303           \\ \hline
\multirow{6}{*}{506 (~33\%)} & Linear Regression    & 0.0505                  & 0.0398           \\ 
                            & Ridge                & 0.0187                  & 0.0198           \\ 
                            & Lasso                & 0.0186                  & 0.0192           \\ 
                            & SVR                  & 0.0208                  & 0.0201           \\ 
                            & XGBRegressor         & 0.023                   & 0.0269           \\ 
                            & KNN                  & 0.0294                  & 0.0299           \\ \hline
\multirow{6}{*}{1536 (~100\%)} & Linear Regression  & 0.0404                  & 0.051            \\ 
                            & Ridge                & 0.0187                  & 0.0199           \\ 
                            & Lasso                & 0.0191                  & 0.0171           \\ 
                            & SVR                  & 0.0203                  & 0.0195           \\ 
                            & XGBRegressor         & 0.0218                  & 0.0199           \\ 
                            & KNN                  & 0.0295                  & 0.029            \\ \hline
\end{tabular}
\label{tab:ada_embedding_results}
\end{table}

\begin{table}[H]
\centering
\caption{MLP Performance for Ada Embedding (50 Repetitions)}
\label{tab:ada_mlp}

\begin{tabular}{|c|c|c|c|}
\hline
\textbf{Dimension (\%)} & \textbf{Metric} & \textbf{Mean (MSE)} & \textbf{Confidence Interval} \\ \hline
76 (~5\%)  & Validation & 0.0169 & [0.0154, 0.0180] \\ 
           & Test       & 0.0218 & [0.0191, 0.0257] \\ \hline
153 (~10\%) & Validation & 0.0191 & [0.0150, 0.0739] \\ 
            & Test       & 0.0211 & [0.0169, 0.0681] \\ \hline
384 (~25\%) & Validation & 0.0157 & [0.0147, 0.0167] \\ 
            & Test       & 0.0187 & [0.0164, 0.0217] \\ \hline
506 (~33\%) & Validation & 0.0154 & [0.0147, 0.0159] \\ 
            & Test       & 0.0181 & [0.0164, 0.0199] \\ \hline
1536 (~100\%) & Validation & 0.0151 & [0.0144, 0.0157] \\ 
              & Test       & 0.0175 & [0.0164, 0.0190] \\ \hline
\end{tabular}
\label{tab:mlp_ada_performance}
\end{table}

\begin{table}[H]
\centering
\caption{MLP Performance for Large Embedding (50 Repetitions)}
\label{tab:larg_mlp}
\begin{tabular}{|c|c|c|c|}
\hline
\textbf{Dimension (\%)} & \textbf{Metric} & \textbf{Mean (MSE)} & \textbf{Confidence Interval} \\ \hline
153 (~5\%)  & Validation & 0.0154 & [0.0139, 0.0164] \\ 
            & Test       & 0.0160 & [0.0137, 0.0189] \\ \hline
307 (~10\%) & Validation & 0.0148 & [0.0140, 0.0155] \\ 
            & Test       & 0.0142 & [0.0127, 0.0158] \\ \hline
768 (~25\%) & Validation & 0.0146 & [0.0136, 0.0155] \\ 
            & Test       & 0.0146 & [0.0129, 0.0164] \\ \hline
1000 (~33\%) & Validation & 0.0144 & [0.0138, 0.0150] \\ 
             & Test       & 0.0138 & [0.0126, 0.0154] \\ \hline
3072 (~100\%) & Validation & 0.0151 & [0.0145, 0.0157] \\ 
              & Test       & 0.0151 & [0.0137, 0.0163] \\ \hline
\end{tabular}
\label{tab:mlp_large_performance}
\end{table}

\section{LLM as Subject}

\subsection{Prompts}

\begin{figure}[H]
    \centering
    \fbox{
        \begin{minipage}{0.9\textwidth}
            \textbf{Instruction:} \\
            Behave like a [Personality Descriptions] \\

            Given the following options, please make a choice for each problem and return only your choices in the format specified. \\

            \textbf{Format:} \\
            (Problem ID, Choice) $|$ (Problem ID, Choice) $|$ ... \\
                
        \end{minipage}
    }
    \caption{Binary Choice Prompt}
    \label{fig:binary_prompt}
\end{figure}

\begin{figure}[H]
    \centering
    \fbox{
        \begin{minipage}{0.9\textwidth}
            \textbf{Instruction:} \\
            Behave like a [Personality Descriptions] \\

            Given the following options, please make a choice for each problem and decide what is your confidence (between 0 to 100) in your choice. Return only your choices and confidence in the format specified. \\

            \textbf{Format:} \\
            (Problem ID, Choice, Confidence) $|$ ... \\
                
        \end{minipage}
    }
    \caption{Confidence Choice Prompt}
    \label{fig:confidence_prompt}
\end{figure}

\begin{figure}[H]
    \centering
    \fbox{
        \begin{minipage}{0.9\textwidth}
            \textbf{Instruction:} \\
            Behave like a [Personality Descriptions] \\

            Given the following options, please indicate your preference for each problem as a percentage, where 0\% represents a complete preference for Option B and 100\% represents a complete preference for Option A. Return your choices in the format specified. \\

            \textbf{Format:} \\
            (Problem ID, Preference) $|$ (Problem ID, Preference) $|$ ...

        \end{minipage}
    }
    \caption{Percentage Choice Prompt}
    \label{fig:percentage_prompt}
\end{figure}

\subsection{Personalities}

\begin{longtable}{>{\RaggedRight}p{3.5cm} >{\RaggedRight}p{5cm} >{\RaggedRight}p{7cm}}
\caption{Decision Making Personalities and Their Characteristics} \\
\toprule
\textbf{Personality} & \textbf{Element} & \textbf{Description} \\
\midrule
\endfirsthead

\toprule
\textbf{Personality} & \textbf{Element} & \textbf{Description} \\
\midrule
\endhead

\midrule
\multicolumn{3}{r}{\textit{Continued on next page}} \\
\midrule
\endfoot

\bottomrule
\endlastfoot

The Calculator & High Sensitivity to Expected Returns & Bases decisions on meticulous calculation of expected outcomes. \\
\hline
The Pessimist & Pessimism & Makes conservative choices to avoid losses, influenced by a negative outlook. \\
\hline
The Equalizer & Bias Toward Equal Weighting & Values simplicity and fairness, treats all information equally. \\
\hline
The Guardian & Sensitivity to Payoff Sign & Sensitive to gains vs. losses, impacting risk assessment. \\
\hline
The Regret Averter & Effort to Minimize Immediate Regret & Focuses on avoiding decisions that might cause regret. \\
\hline
The Adaptive & Impact of Feedback on Sensitivity to Probability & Changes decision-making strategy based on feedback and probability updates. \\
\hline
The Analyst & Various BEAST Elements & Uses a methodical approach, reviews data, considers multiple perspectives. \\
\hline
The Realist & Pragmatic Assessment & Makes decisions based on pragmatic assessment of available options. \\
\hline
The Optimist & Expecting Favorable Outcomes & Sees potential for positive outcomes, more likely to take risks. \\
\hline
The Minimalist & Simplicity in Decisions & Prefers simplicity, choosing the simplest option available. \\
\hline
\label{table:beast_persona}

\end{longtable}

\subsection{Training Details}

For the baseline condition, the \textbf{Binary} condition performed best. The best regressor was the \textbf{Support Vector Regressor (SVR)} with the following hyperparameters: 
\textbf{C = 0.1}, \textbf{epsilon = 0.1}, \textbf{gamma = "auto"}, and \textbf{kernel = "rbf"}.

For the personalities condition, the \textbf{Confidence} condition performed best. The best regressor was the \textbf{Random Forest Regressor} with the following hyperparameters: 
\textbf{n\_estimators = 300}, \textbf{max\_depth = 4}, \textbf{min\_child\_weight = 6}, \textbf{learning\_rate = 0.05}, \textbf{reg\_lambda = 10}, and \textbf{reg\_alpha = 0.5}.

\subsection{Results}

Table \ref{tab:all_conditions_combined} presents results of all the conditions.

\begin{table}[h!]
\centering
\begin{tabular}{lcccc}
\toprule
\textbf{Condition} & \textbf{Model} & \textbf{CV Score} & \textbf{Test MSE} \\
\midrule
\multirow{6}{*}{\textbf{Binary}} & LinearRegression & 0.0185 & 0.0178 \\
 & Ridge & 0.0185 & 0.0178 \\
 & SVR & 0.0181 & 0.0172 \\
 & KNeighborsRegressor & 0.0184 & 0.0192 \\
 & RandomForestRegressor & 0.0208 & 0.0222 \\
 & XGBRegressor & 0.0184 & 0.0193 \\
\midrule
\multirow{5}{*}{\textbf{Percentage}} & Ridge & 0.0298 & 0.0294 \\
 & SVR & 0.0288 & 0.0283 \\
 & KNeighborsRegressor & 0.0296 & 0.0291 \\
 & RandomForestRegressor & 0.0302 & 0.0297 \\
 & XGBRegressor & 0.0288 & 0.0287 \\
\midrule
\multirow{6}{*}{\textbf{Confidence}} & LinearRegression & 0.0198 & 0.0238 \\
 & Ridge & 0.0198 & 0.0238 \\
 & SVR & 0.0197 & 0.0233 \\
 & KNeighborsRegressor & 0.0204 & 0.0257 \\
 & RandomForestRegressor & 0.0215 & 0.0300 \\
 & XGBRegressor & 0.0199 & 0.0244 \\
\midrule
\multirow{6}{*}{\textbf{Combined}} & LinearRegression & 0.0172 & 0.0177 \\
 & Ridge & 0.0172 & 0.0177 \\
 & SVR & 0.0174 & 0.0170 \\
 & KNeighborsRegressor & 0.0174 & 0.0175 \\
 & RandomForestRegressor & 0.0209 & 0.0211 \\
 & XGBRegressor & 0.0173 & 0.0178 \\
\bottomrule
\end{tabular}
\caption{Fit results for the Binary, Percentage, Confidence, and Combined conditions across various models.}
\label{tab:all_conditions_combined}
\end{table}

\newpage

\section{Feature Extraction}

\subsubsection{Prompts}

\begin{longtable}{|p{3cm}|p{12cm}|}
\caption{Different Decision-Making Prompt Types and Their Instructions} \label{tab:decision_prompts} \\
\hline
\textbf{Prompt Type} & \textbf{Instruction} \\
\hline
\endfirsthead

\hline
\textbf{Feature} & \textbf{Prompt} \\
\hline
\endhead

\hline
\multicolumn{2}{r}{{Continued on next page}} \\
\hline
\endfoot

\hline
\endlastfoot

Unbiased & Given two options: \newline Option A: \{A\} \newline Option B: \{B\} \newline Let's say I simulate these options several times. In each round, I draw one outcome from each option and check which option provided the better (higher) payoff, if any. Can you assess which option yields more rounds with a strictly better payoff? If it is too hard to tell, say so. \newline Take your time, analyze, think it thoroughly, and then only provide a final answer without explanations. \\

\hline
Sign & Given two options: \newline Option A: \{A\} \newline Option B: \{B\} \newline Let's say I simulate these options several times. In each round, I draw one outcome from each option and record the outputs. Then, I sign-transform all of these outcomes and check, in each round, which option provided the better payoff-sign (ignoring the payoff size), if any. Can you assess which option yields more rounds with a strictly better payoff sign? If it is too hard to tell, say so. \newline Take your time, analyze, think it thoroughly, and then only provide a final answer without explanations. \\

\hline
Better on Avg & Given two options: \newline Option A: \{A\} \newline Option B: \{B\} \newline Let's say I simulate these options several times. In each round, I draw one outcome from each option and record the outputs. Then, for each option, I sum the payoffs each option yielded across all rounds. Can you assess which option yields a higher sum of payoffs, if any? If it is too hard to tell, say so. \newline Take your time, analyze, think it thoroughly, and then only provide a final answer without explanations. \\

\hline
Uniform & Given two options: \newline Option A: \{A\} \newline Option B: \{B\} \newline Let's say I simulate these options several times. In each round, I first transform all payoffs in each option to be equally likely and then draw one outcome. That is, I transform each option's payoff distribution so that actual probabilities are ignored, and all its payoffs have the same probability to be drawn before I make draws from these transformed distributions. Then, I record the outputs and check, in each round, which option provided the better payoff, if any. Can you assess which option yields more rounds with a strictly better payoff under this transformation? If it is too hard to tell, say so. \newline Take your time, analyze, think it thoroughly, and then only provide a final answer without explanations. \\

\hline
Dominance (Dom) & Given two options: \newline Option A: \{A\} \newline Option B: \{B\} \newline Let's say I simulate these options several times. In each round, I draw one outcome from each option and check which option provided the better (higher) payoff, if any. Can you assess *if* one option yields a payoff that is at least as good as the other option payoff across *all* rounds? If this is not the case, please clearly state that by answering 'No'. If it is too hard to tell, say so. \newline Take your time, analyze, think it thoroughly, and then only provide a final answer without explanations. \\

\hline
Worst Case & Given two options: \newline Option A: \{A\} \newline Option B: \{B\} \newline Let's say I simulate these each of these options once, and each option yields its worst (lowest) payoff. Can you assess which option, if any, yields a better payoff in this scenario? If it is too hard to tell, say so. \newline Take your time, analyze, think it thoroughly, and then only provide a final answer without explanations. \\

\hline
Risk & Given two options: \newline Option A: \{A\} \newline Option B: \{B\} \newline Can you assess which option, if any, is riskier (i.e., has higher variance)? If it is too hard to tell, say so. \newline Take your time, analyze, think it thoroughly, and then only provide a final answer without explanations. \\

\hline
\caption{Prompt for Extracting Features}
\label{ddd}
\end{longtable}

\subsubsection{Training Details}

For the feature extraction regressor, \textbf{XGBoost} was used with the following hyperparameters:  

\begin{itemize}
    \item \textbf{Subsample:} 0.6
    \item \textbf{Scale Pos Weight:} 100
    \item \textbf{Regularization Lambda ($\lambda$):} 0.01
    \item \textbf{Regularization Alpha ($\alpha$):} 0.01
    \item \textbf{Number of Estimators:} 100
    \item \textbf{Min Child Weight:} 5
    \item \textbf{Max Depth:} 10
    \item \textbf{Max Delta Step:} 0
    \item \textbf{Learning Rate:} 0.05
    \item \textbf{Gamma:} 0
    \item \textbf{Colsample by Tree:} 0.7
    \item \textbf{Colsample by Level:} 0.4
\end{itemize}

\end{document}